\documentclass[letterpaper, 10 pt, conference]{ieeeconf}  

\IEEEoverridecommandlockouts                              

\usepackage{graphicx} 
\graphicspath{{figures/}{./figures/}}

\title{\LARGE \bf
Transfer Learning for Action Unit Recognition
}

\author{Yen Khye Lim$^{1}$, Zukang Liao$^{1}$, Stavros Petridis$^{1}$ and Maja Pantic$^{1, 2}$
\thanks{$^{1}$Faculty of Engineering, Department of Computing, Imperial College London, SW7 7AZ, United Kingdom
        }
\thanks{$^{2}$Faculty of Engineering, Department of Computing, University of Twente, 7500 AE Enschede, The Netherlands
}}

\begin{document}

\maketitle
\thispagestyle{empty}
\pagestyle{empty}

\begin{abstract}

This paper presents a classifier ensemble for Facial Expression Recognition (FER) based on models derived from transfer learning. The main experimentation work is conducted for facial action unit detection using feature extraction and fine-tuning convolutional neural networks (CNNs). Several classifiers for extracted CNN codes such as Linear Discriminant Analysis (LDA), Support Vector Machines (SVMs) and Long Short-Term Memory (LSTM) are compared and evaluated. Multi-model ensembles are also used to further improve the performance. We have found that VGG-Face and ResNet are the relatively optimal pre-trained models for action unit recognition using feature extraction and the ensemble of VGG-Net variants and ResNet achieves the best result.


\end{abstract}

\section{INTRODUCTION}
While deep networks have been successful for general object classification tasks, Facial Expression Recognition (FER) remains a challenging problem. There are many obstacles when implementing deep networks for FER. Firstly, these algorithms typically rely on a very large set of annotated examples. Furthermore, expensive hardware such as high-end Graphics Processing Units (GPUs) is required. Moreover, these networks demand high computational power, with training time ranging from weeks to months. Due to these limitations, transfer learning from publicly available CNNs is proposed to exploit the powerful representations of successful networks.

Most modern classifiers are built around the six universal emotions i.e. anger, disgust, fear, happiness, sadness and surprise \cite{ekman1971constants}, alongside the neutral expression as an optional class. Another approach for FER involves the use of the Facial Action Coding System (FACS) which describes facial muscle movement using 44 different Action Units (AU) \cite{princefacial}. Each AU corresponds to a specific facial substructure and the six general emotions can be categorised using a combination of multiple AUs. The Emotional Facial Action Coding System (EMFACS) is a subset of FACS which considers only the relevant AUs responsible for such expressions. Therefore, images can be segmented into different regions according to which AU we are recognising and then segmented regions can be fed into CNNs to improve performance. This project focuses on recognising occurrence of 12 AUs using transfer learning. The results, i.e. combinations of occurrence of different AUs, can then be further used for other facial recognition tasks including FER.

When only one network is used to do feature extraction for action unit recognition, we have found that the best performance comes from VGGNet and ResNet. In addition, fine-tuning CNNs, using a region-based approach and using an ensemble can all further improve the performance.


\section{CNN Architectures}\label{s_pop_arch}

In order to explore which CNN architecture performs the best for action units recognition, we have used LeNet-5 \cite{lecun1998gradient}, AlexNet \cite{krizhevsky2012imagenet}, ZFNet \cite{zeiler2014visualizing}, VGGNet \cite{Simonyan14c}, GoogleNet \cite{szegedy2014going} and ResNet \cite{he2016deep} to conduct experiments.

\subsection{AlexNet}

In 2012, AlexNet by Krizhevsky topped the ImageNet Large Scale Visual Recognition Challenge (ILSVRC) by a significant lead \cite{krizhevsky2012imagenet}. It managed to reduce the top 5 error rate from 26\% to 16\%. The architecture was trained on multiple GPUs for efficient implementation of convolution operations. AlexNet is a large network with input size 227$\times$227$\times$3 pixels. It introduced stacking of convolutional layers before pooling. Furthermore, AlexNet is the first architecture to use ReLU non-linearity and normalisation layers. It consists of five convolutional layers and three fully-connected layers. Finally, a 1000-way softmax calculates the distribution over all the class labels.

\subsection{ZFNet}

ZFNet, the champion of ILSVRC 2013 modified some hyperparameters for the middle convolutional layers of AlexNet \cite{zeiler2014visualizing}. In particular, the first convolutional layers was changed from 11$\times$11 with stride 4 to 7$\times$7 with stride 2. On the last three convolutional layers, the number of filters were increased from 384, 384, 256 to 512, 1024, 512 respectively.

\subsection{VGGNet}

The runner-up in ILSVRC 2014 is VGGNet \cite{Simonyan14c} which features a consistent architecture while increasing the number of layers. VGGNet used only 3$\times$3 convolutions with one stride and zero-padding. For all pooling layers, 2$\times$2 max pooling of stride 2 without zero-padding are implemented. For VGG-16, there are many variants including VGG-Face  \cite{parkhi2015deep}, VGG-Fast, VGG-Medium and VGG-Slow \cite{Chatfield14}. Particularly, VGG-Face was trained on 2.6 million images to classify 2622 different individuals while the original task for VGGNet was object recognition. For VGG-Fast, VGG-Medium and VGG-Slow, the faster speed mainly results from using larger strides and fewer channels, which leads to less connectivity between convolutional layers. The sturctures of VGG-Fast, VGG-Medium and VGG-Slow are shown in Table~\ref{t_net_vgg_f}.

\begin{table}[!h]
	\caption{Network configuration of VGG-Fast/CNN-F, VGG-Medium/CNN-M and VGG-Slow/CNN-S \cite{Chatfield14}}
	\includegraphics[scale=0.19]{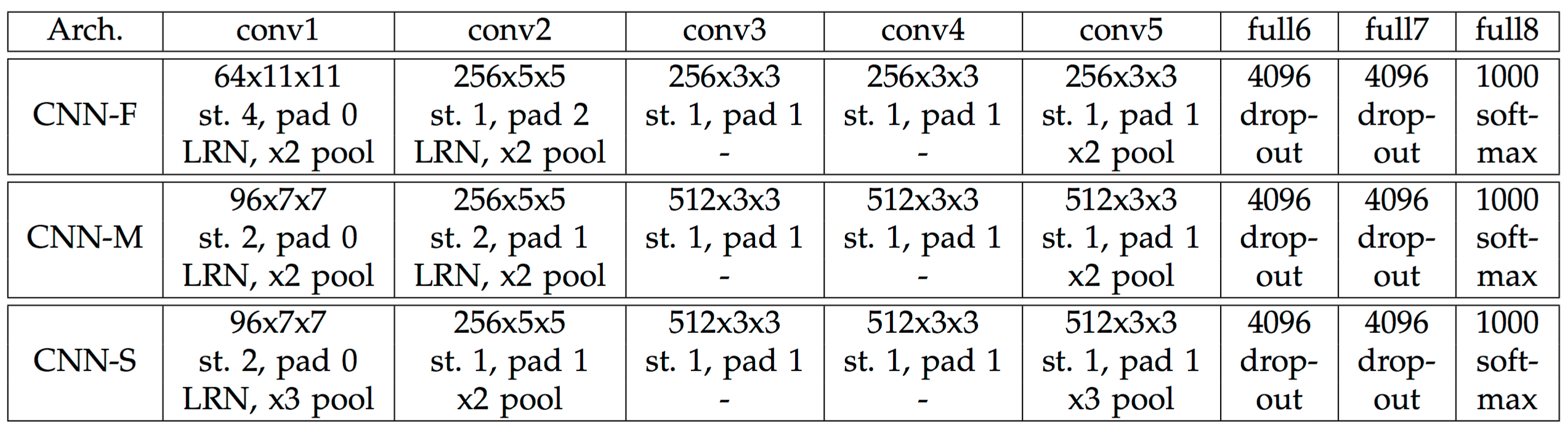}
	\label{t_net_vgg_f}
\end{table}

\subsection{GoogLeNet}

GoogLeNet which won ILSVRC 2014 with top 5 error of 6.7\% introduced an ``Inception" module which optimises the number of parameters in the network \cite{szegedy2014going}. GoogLeNet has a complex architecture of 22 layers. It also completely removes the fully-connected layers in favour of average pooling. 


\subsection{ResNet}
ResNet by Microsoft Research won the ILSVRC 2015 competition with a stunning top 5 error rate of 3.57$\%$ \cite{he2016deep}. They demonstrated state-of-the-art results by experimenting with networks of different depths, from 18 layers up to 152 layers as shown in Table~\ref{t_resnet}. ResNet adds a batch normalisation layer after every convolutional layer and omits dropout layers. This architecture explores a new method of constructing deeper networks by using a residual network structure as opposed to traditional plain networks. By adding identity ``shortcut connections" $x$ in a residual network, the new layer is able to learn something different apart from those which have been already encoded in the previous layer.

\begin{table}[!h]
	\centering
	\caption{Architecture of ResNet \cite{he2016deep}}
	\includegraphics[scale=0.1]{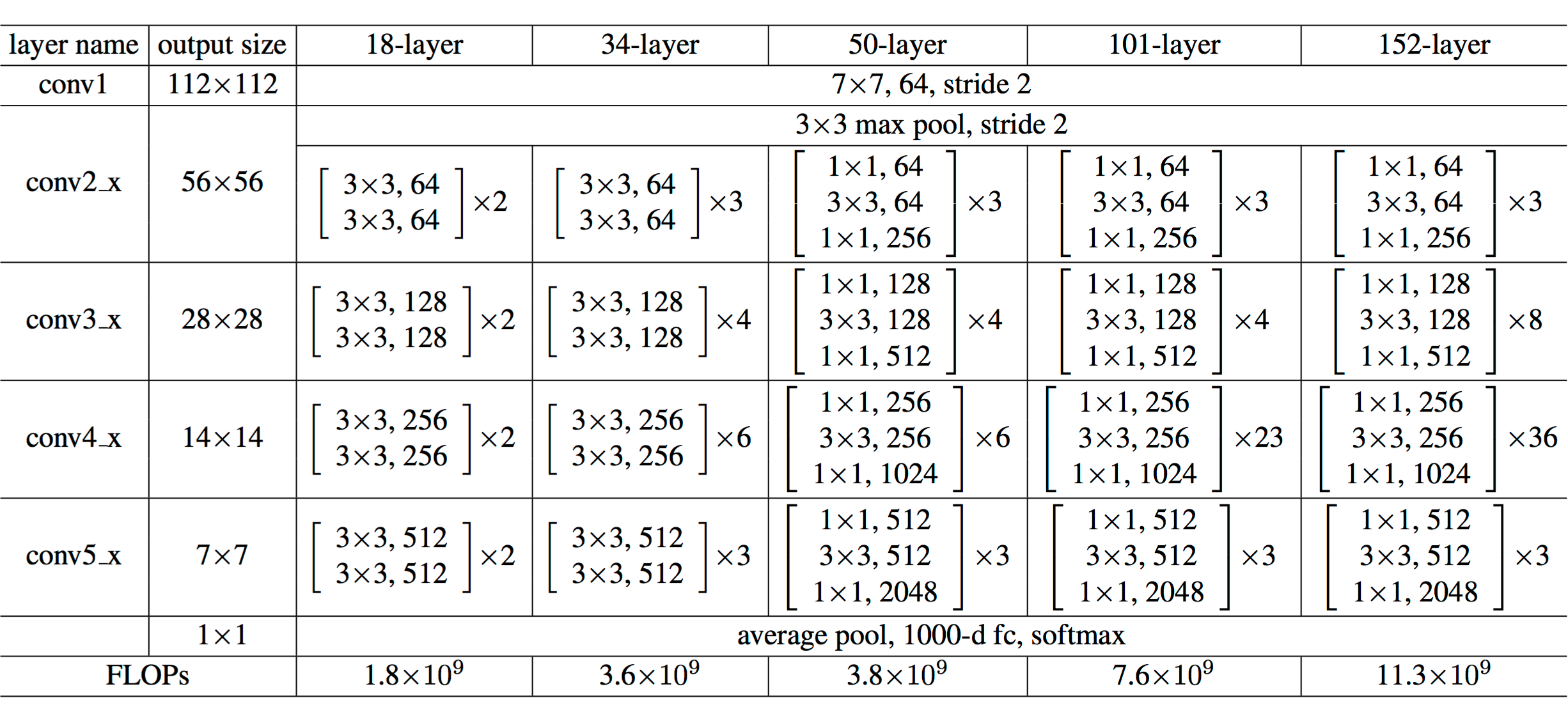}
	\label{t_resnet}
\end{table}

\section{Database}

\begin{figure}[!h]
	\centering
	\includegraphics[scale=0.27]{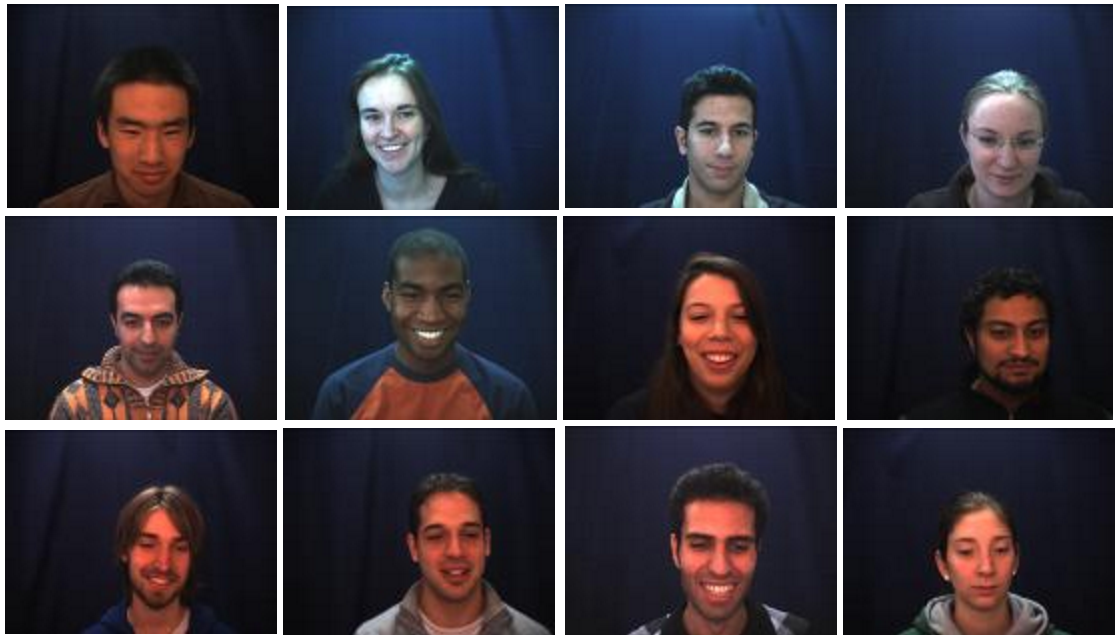}
	\caption{Sample images from the DISFA database}
	\label{f_disfa_sample}
\end{figure}

The Denver Intensity of Spontaneous Facial Action (DISFA) database consists of non-posed videos where subjects show different facial expressions as shown in Figure~\ref{f_disfa_sample}. The videos are labelled both the occurrence and intensity values for 12 AUs as shown in Table~\ref{f_disfa_facs} by FACS \cite{mavadati2013disfa}. Among the 27 subjects of various ethnicity, 12 are females and 15 are males. The stereo videos of length 4845 frames for each subject are recorded at 1024$\times$768 resolution. Moreover, this database includes 68 facial landmark points for all images.

\begin{table}[!h]
	\centering
	\caption{Action Unit labels provided in the DIFSA database \cite{ekman1978facial}}
	\label{f_disfa_facs}
	\includegraphics[scale=0.42]{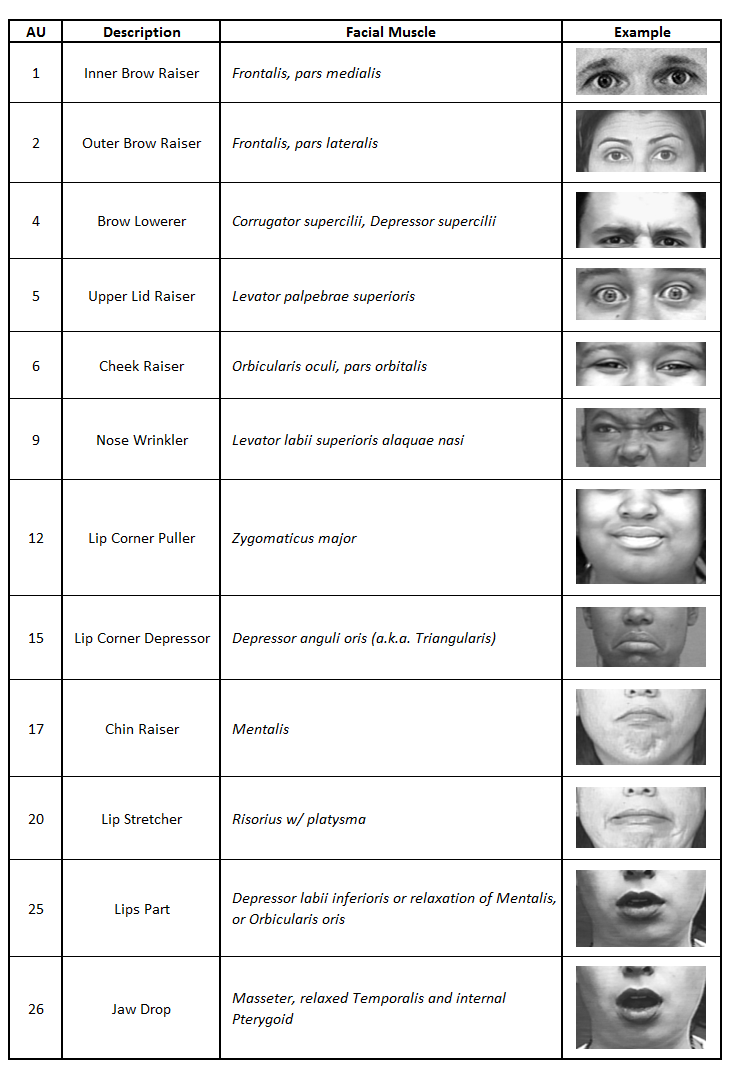}
\end{table}

The training, validation and testing sets consist of 12, 6 and 9 subjects respectively. The intensity values are provided for each frame where 0 indicates the absence of an AU while 5 is the maximum intensity. For this experimentation, a threshold is set at equal or larger than 2 for an AU to be considered present. Among the total 130814 frames, the distribution of the AU activations is tabulated in Table \ref{t_disfa_dist}.

\begin{table}[h]
\centering
\caption{Number of frames with AU activations in DISFA database}
\label{t_disfa_dist}
\begin{tabular}{ccccccc}
\hline
AU  & 1     & 2    & 4     & 5    & 6     & 9     \\\hline
No. & 6506  & 5644 & 19933 & 1150 & 10327 & 5473  \\\hline
\\\hline
AU  & 12    & 15   & 17    & 20   & 25    & 26    \\\hline
No. & 16851 & 2682 & 6588  & 2941 & 36247 & 11533 \\\hline
\end{tabular}
\end{table}

\section{Feature extraction}
\begin{figure}[!h]
	\centering
	\includegraphics[scale=0.29]{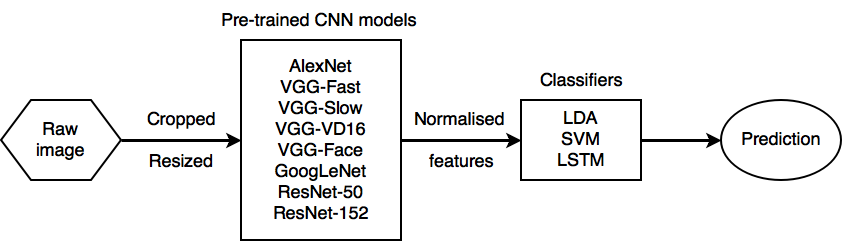}
	\caption{Overview of feature extraction methodology}
	\label{f_disfa_diag}
\end{figure}

CNNs trained on the ImageNet dataset as discussed in Section~\ref{s_pop_arch} are used as feature extractors. The overview of feature extraction methodology is shown in Figure~\ref{f_disfa_diag}. 

\subsection{Pre-processing} \label{sec-Pre-processing}
Due to the large number of negative examples, the training and validation sets are balanced by randomly choosing a smaller subset based on the total number of positive examples. Images are aligned and cropped given the feature points. Also, images are normalised according to which network is used. For example, when VGG-Face is used, the mean image of the training set is subtracted from every single image \cite{parkhi2015deep}.

\subsection{Use CNNs to extract features}
Various publicly available CNNs which have been trained on large datasets can be used as fixed feature extractors. The dimensionality of features extracted from different CNN structures are shown in Table~\ref{Dimensionality_CNNs}. In this project, we extract features from the first fully connected layer after convolutional blocks. For example, if VGGNet variants are used, features from fc6 layer would be extracted.

\begin{table}[h]
\centering
\caption{Dimensionality of extracted features}
\label{Dimensionality_CNNs}
\begin{tabular}{cccccc}
\hline
           & AlexNet & ZFNet & VGGNet & GoogLeNet & ResNet \\\hline
Dimensions & 2048    & 4096  & 4096   & 1024      & 2048  \\\hline
\end{tabular}
\end{table}

\subsection{Training Linear Classifiers}
After features have been extracted, they are normalised by setting the mean and standard deviation to 0 and 1, respectively, based on the training set. If the task of AU recognition is conducted for each frame, these features of 1024 (GoogLeNet) or 2048 (ResNet, AlexNet) or 4096 (VGGNet variants) dimensions are used to train linear classifiers. We treat every single action unit independently, so one classifier model is learnt for each AU. Both Linear Discriminant Analysis (LDA) and Support Vector Machines (SVMs) with linear kernel are trained for this binary classification task.


\subsection{LSTM for Temporal Information}
In order to capture variations in the temporal domain, LSTM models are implemented.

For this model, features from ResNet-152 are used due to the lower dimensionality (i.e. 2048 dimensions) as compared with feature vectors extracted from VGG variants (i.e. 4096 dimensions). The LSTM input layer accepts 2048-dimensional arrays and has a single hidden layer of 200 units. The output layer is a binary classification layer which determines if an AU is present or absent. The learning rate is set to 0.0001 and the momentum is fixed at 0.9. To provide robustness against overfitting, Gaussian noise with a zero mean and a standard deviation of 0.1 is applied to the weights for each batch before computing the gradient.

For each subject, the 4845 frames are divided into smaller sequences based on the activations of AU. Additionally, three frames are added before and after each string of activation so that the model can learn these transitions. To cater for the imbalance in this dataset, long inactive sequences ranging from 500 to 1000 frames in the training and validation set are filtered and removed. However, the testing set is treated as a full sequence of 4845 length for each subject.

\subsection{Region-based Approach}\label{region_based_approach}
\begin{figure}[!h]
	\centering
	\includegraphics[scale=0.3]{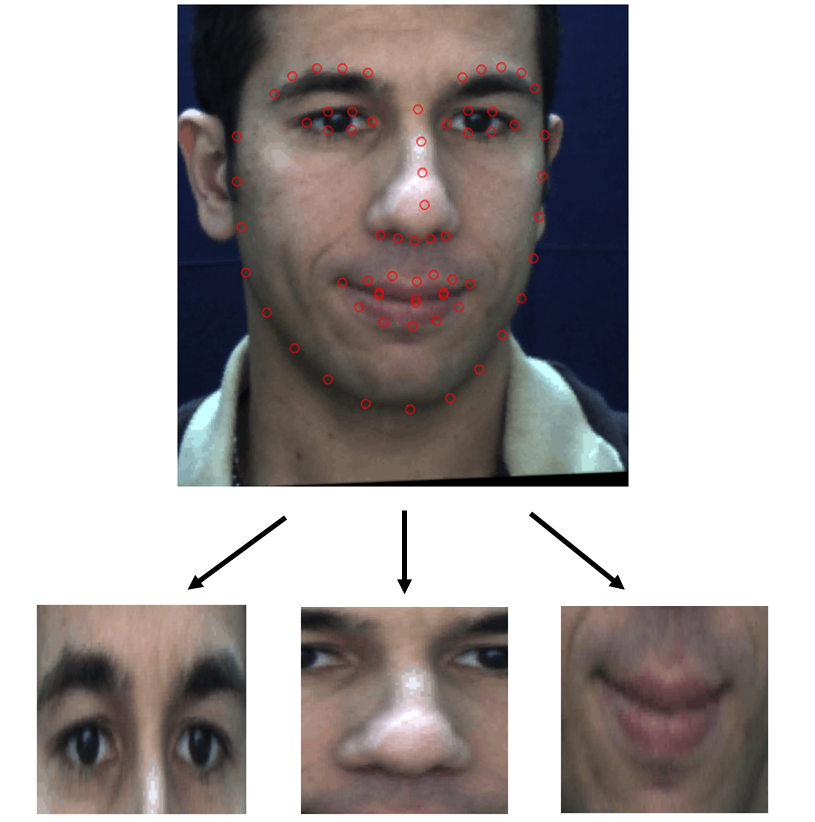}
	\caption{Segmentation based on facial landmark points into upper half, middle and lower half regions}
	\label{f_seg}
\end{figure}

Instead of using the entire face for training, a more concise approach would be to segment the original face based on the region of activation for each AU. The region of the face that each AU corresponds to is shown in Table \ref{t_disfa_seg}. Since each classifier is unique to one action unit, only the information from a specific region is required as input. For this experiment, the original face is split into three distinct regions as shown in Figure~\ref{f_seg}.

\begin{table}[!h]
\centering
\caption{Different regions are used for each action unit}
\begin{tabular}{ccc}\hline
AU & Description          & Region     \\\hline
1  & Inner Brow Raiser    & Upper half \\
2  & Outer Brow Raiser    & Upper half \\
4  & Brow Lowerer          & Upper half \\
5  & Upper Lid Raiser     & Upper half \\
6  & Cheek Raiser         & Upper half \\
9  & Nose Wrinkler        & Middle \\
12  & Lip Corner Puller    & Lower half   \\
15 & Lip Corner Depressor & Lower half  \\
17 & Chin Raiser          & Lower half \\
20 & Lip Stretcher        & Lower half \\
25 & Lips Part            & Lower half \\
26 & Jaw Drop             & Lower half \\\hline
 
\end{tabular}
\label{t_disfa_seg}
\end{table}

\section{fine-tuning}
Another approach of transfer learning is to reuse existing models and fine-tune the network to learn the variations of a new task. In this work, fine-tuning is conducted on the VGG-Fast model due to the limitation of GPUs' memory. We use the same pre-processing method as described in Section~\ref{sec-Pre-processing}. The configuration used for fine-tuning is similar to the original model, except that the weights for all fully connected layers have been reinitialised and the softmax layer is replaced by two neurons (AU is present or absent) to adapt to the task. The motivation behind this configuration is that features from shallower layers should be more generic and transferable to another problem domain, while features from deeper layer (e.g. fully connected layers) are more abstract and task-specific. Therefore, we retain all weights of convolutional layers and reinitialise weights of fully connected layers.




\section{Classifier Ensemble}

While transfer learning itself can provide decent results, a collections of models should be able to push the boundaries of performance. There are several methods available for combining multiple models to produce an output. For this experiment, a simple approach of using majority voting system is implemented. Ideally, one model from each family or variant is selected as different architectures learn unique representations of the input data.
 

The following three classifier ensembles are evaluated:
\begin{enumerate}
	\item Models from feature extraction only.
	\item Models from feature extraction and fine-tuned CNNs.
	\item Models from feature extraction and fine-tuned CNNs and region-based approach.
\end{enumerate}

\section{Results}
F$_1$ score is used as the primary criteria alongside classification rate due to the large imbalance in the DISFA dataset. Moreover, as we are exploring which CNN structure performs the best for these 12 action units in the DISFA databset, the f1 score and classification rate of each action units are averaged. The averaged f1 score and classification rate are the final metrics to evaluate a certain pre-trained network or a classifier ensemble.

\subsection{Feature extraction results}
\begin{figure}[!h]
	\centering
	\includegraphics[scale=0.26]{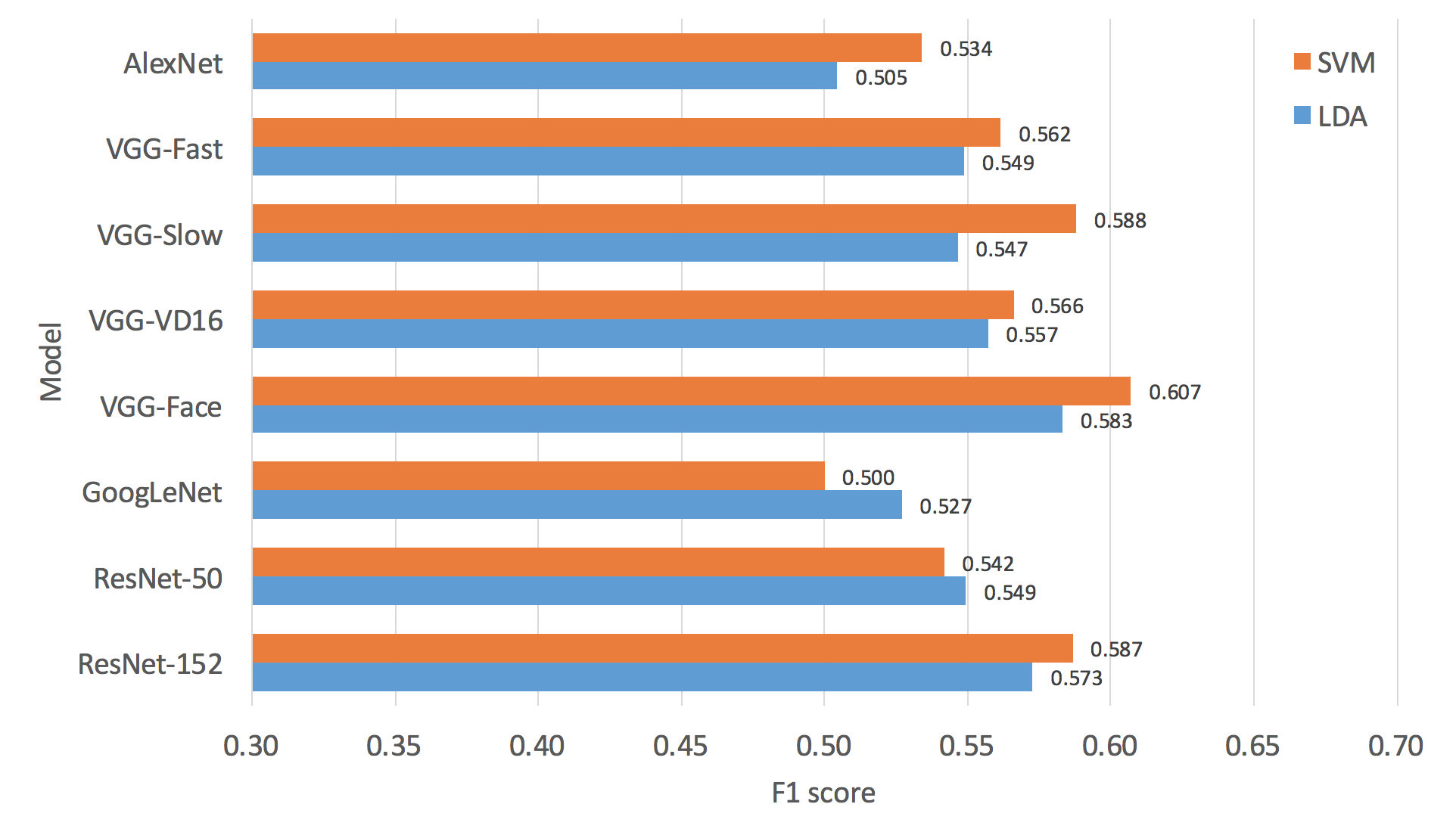}
	\caption{F$_1$ score of feature extraction from several pre-trained CNNs using various LDA and SVM classifiers}
	\label{f_bar_lda_svm}
\end{figure}

The average performance of each pre-trained CNN model for both LDA and SVM classifiers is summarised in Figure \ref{f_bar_lda_svm}. The performance of SVM classifiers mimics the trend of LDA models. SVMs work particularly well with high dimensional inputs. Overall, the results obtained using SVMs are slightly better than LDA classifiers. The best results of feature extraction are from VGG-Face and ResNet-152. They achieve an F$_1$ score of 60.7\% and 58.7\% when an SVM is used. 


\begin{figure}[!h]
	\centering
	\includegraphics[scale=0.34]{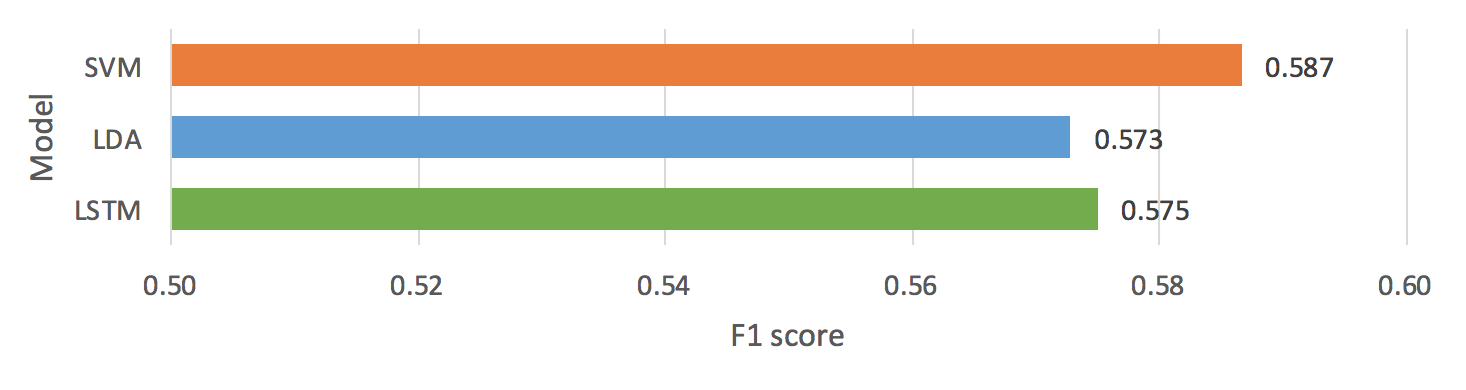}
	\caption{F$_1$ score of various classifiers using ResNet-152 features}
	\label{f_bar_lstm}
\end{figure}

The performance of the LSTM model trained using ResNet-152 features is compared against the LDA and SVM classifiers in Figure \ref{f_bar_lstm}. The classifier is able to learn temporal features in the input image sequences. However, LSTMs do not scale well when the input dimensions are large, e.g. 2048 or 4096 dimensions. Moreover, LSTM models require a series of images sequences unlike LDA and SVMs which are trained on each frame. Due to the small number of training sequences and large dimensionality of the input, the LSTM did not perform as well as the SVM counterpart. However, this model can still be useful as it encodes the time variations of input features.

\subsection{Fine-tuning results}
\begin{figure}[!h]
	\centering
	\includegraphics[scale=0.23]{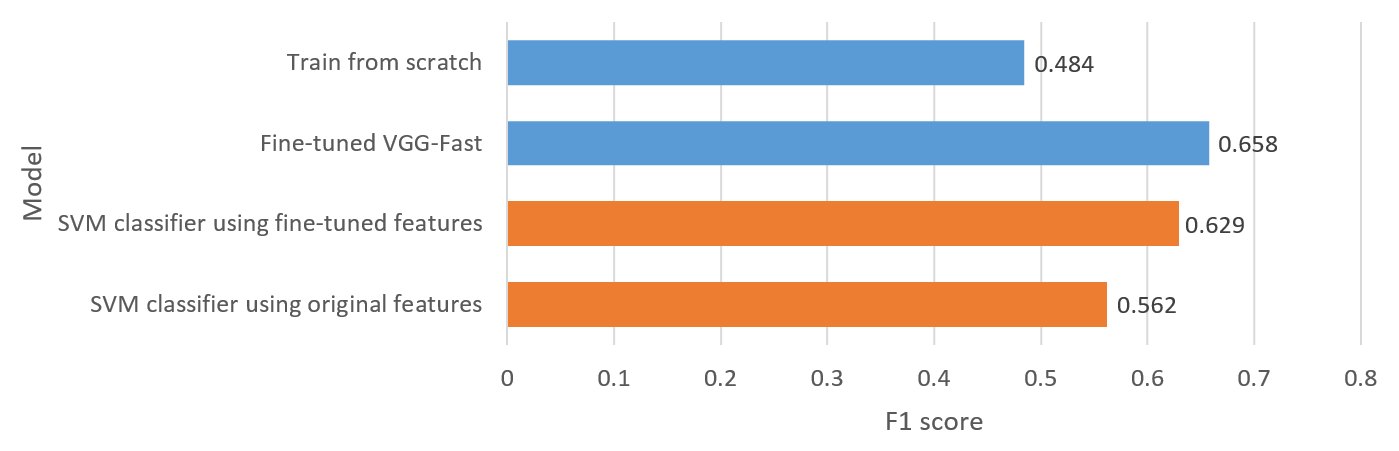}
	\caption{F$_1$ score of different models related to fine-tuning for DISFA}
	\label{f_bar_ft_disfa}
\end{figure}

Fine-tuning VGG-Fast achieves an F$_1$ score of 65.8\% which is 5.1\% better than the best results obtained from feature extraction as shown in Figure \ref{f_bar_ft_disfa}. To further investigate whether the features learnt in a fine-tuned network are actually better than the original features that can be extracted from the same layer, features from the fc6 layer of VGG-Fast are extracted before and after fine-tuning to be compared using SVM classifiers. Using the features of a fine-tuned model, the F$_1$ score of the SVM models increased from 56.2\% to 62.9\%. Therefore, fine-tuning pre-trained CNN models can improve the performance and make the middle-layer features more suitable for a specific task.

\subsection{Region-based approach results}
\begin{figure}[!h]
	\centering
	\includegraphics[scale=0.24]{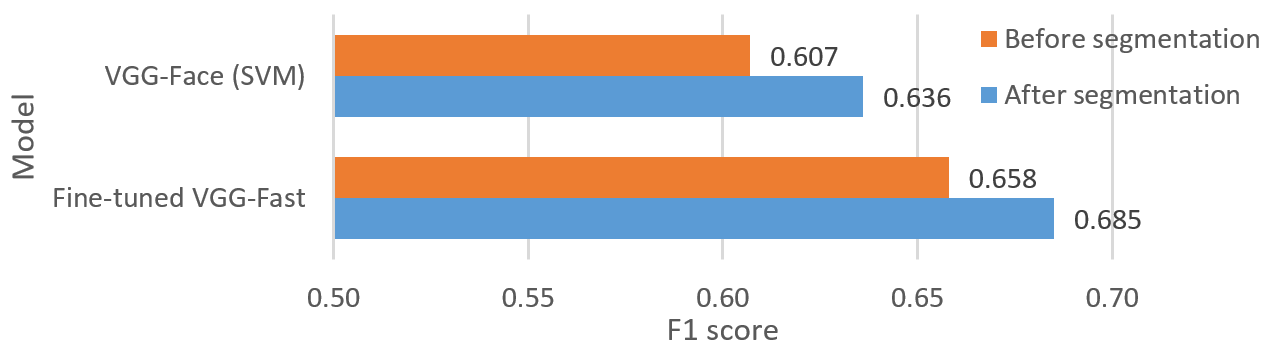}
	\caption{F$_1$ score of different models before and after face segmentation}
	\label{f_bar_seg}
\end{figure}

When using AU-specific image regions as inputs, both feature extraction and fine-tuning result in slightly better performance as illustrated in Figure \ref{f_bar_seg}. When comparing with SVM classifiers trained on VGG-Face features, a 2.9\% increase in F$_1$ score is observed. Fine-tuning VGG-Fast with FC6 and FC7 weights reinitialised resulted in the best performing model with 68.5\% F$_1$ score. 


\subsection{Ensemble results}
\begin{figure}[!h]
	\centering
	\includegraphics[scale=0.24]{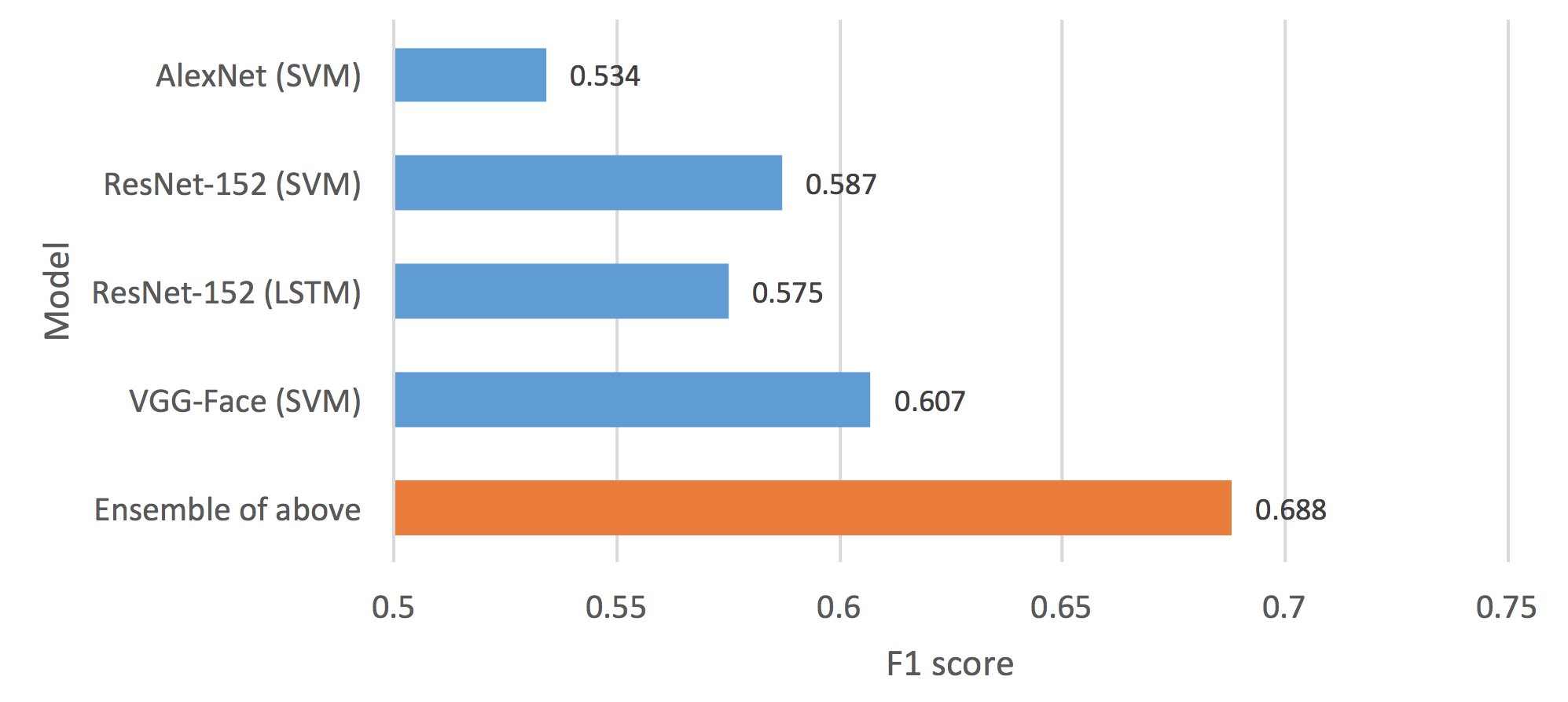}
	\caption{F$_1$ score of classifier ensemble using models from feature extraction only}
	\label{f_ens_fe}
\end{figure}

An ensemble using majority voting is implemented in this work. Besides the tweaked GooLeNet model, one classifier from each variant is selected to form an ensemble. Figure \ref{f_ens_fe} illustrates the performance of this approach by taking a group of SVM and LSTM classifiers. The ensemble method results in 68.8\% F$_1$ score, which is superior than the results of fine-tuning VGG-Fast. This shows that using a collection of models from feature extraction is a feasible approach if fine-tuning is not an option.

The second ensemble combines models from both feature extraction and fine-tuned CNNs. Since AlexNet (SVM) has the poorest performance, it is replaced by the fine-tuned VGG-Fast model. As a result, the F$_1$ score increased by 2.3\% to 71.1\%.

\begin{figure}[!h]
	\centering
	\includegraphics[scale=0.25]{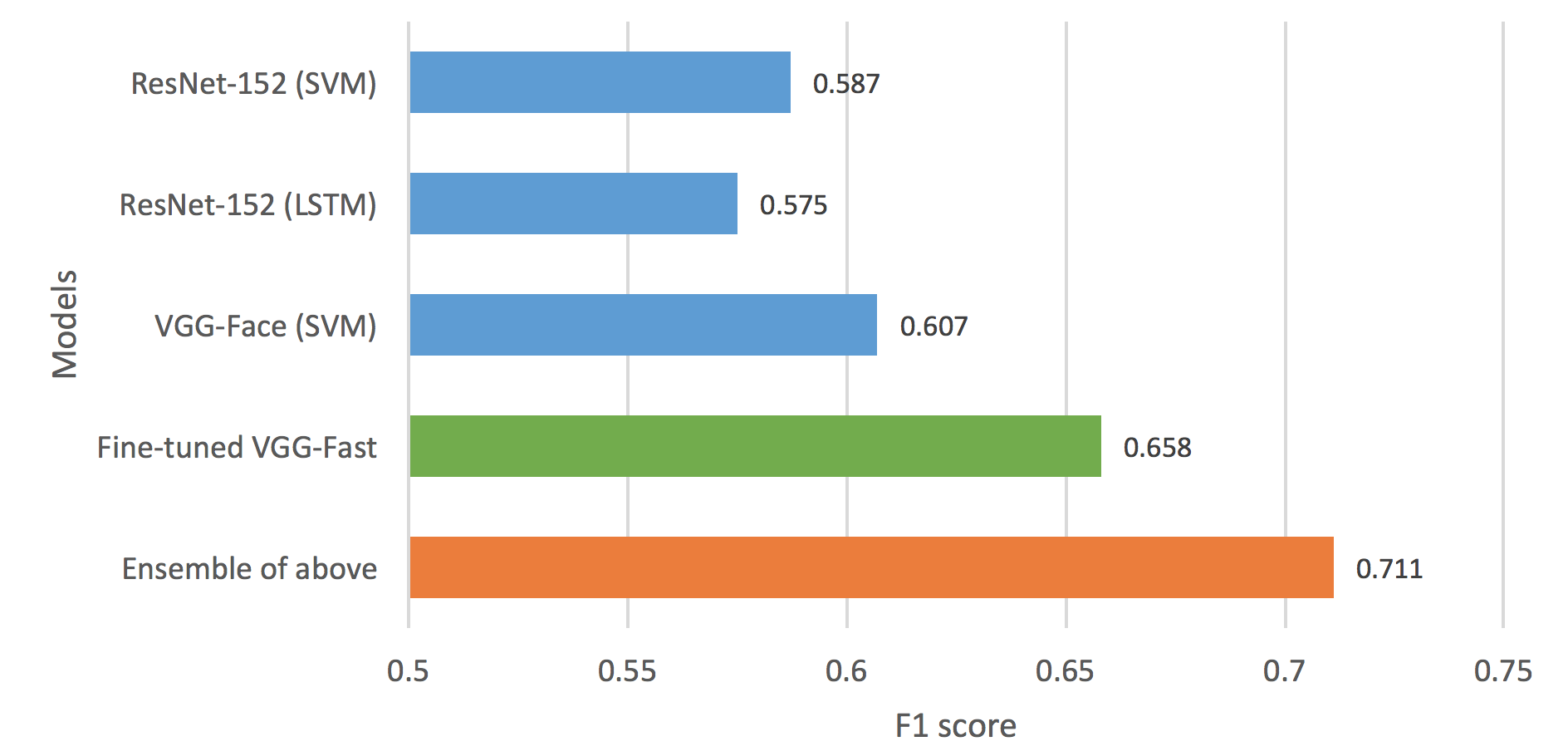}
	\caption{F$_1$ score of classifier ensemble using models from feature extraction and fine-tuned CNNs}
	\label{f_ens_feft}
\end{figure}

The best possible result is achieved when we add to the ensemble models trained on face regions as described in Section~\ref{region_based_approach}. An F$_1$ score of 74.1\% is reported which is 5.6\% higher than the best individual model of fine-tuned VGG-Fast.


\begin{figure}[!h]
	\centering
	\includegraphics[scale=0.32]{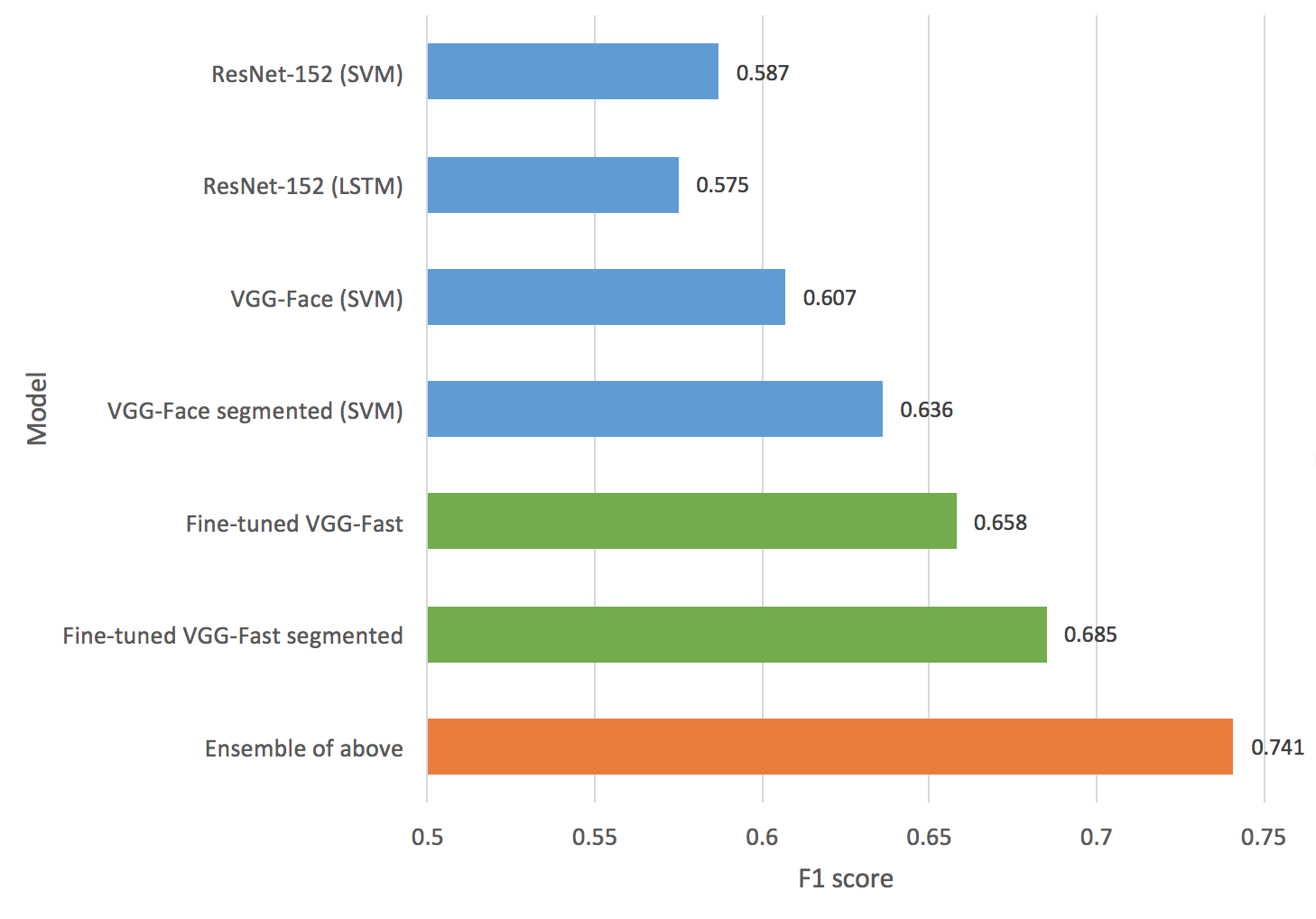}
	\caption{F$_1$ score of classifier ensemble using models from feature extraction and fine-tuned CNNs with segmented faces}
	\label{f_ens_all}
\end{figure}

\section{CONCLUSIONS AND FUTURE WORK}
This project focuses on occurrence detection of each AU among the 12 AUs in the DISFA dataset using transfer learning. Feature extraction is conducted using a wide variety of models including AlexNet, VGG-Fast, VGG-Slow, VGG-VeryDeep16, VGG-Face, GoogLeNet, ResNet-50 and ResNet-152. Based on these CNN codes, LDA, SVM and LSTM classifiers are trained. As a result, SVM classifiers tend to perform better than their LDA counterparts while their LSTM counterparts performed the worst due to the limited number of positive examples in the DISFA dataset. For feature extraction, VGG-Face and ResNet produce the best results. We have also shown that fine-tuning CNNs, using a region-based approach and using an ensemble approach can produce better results.

A proposed extension of this project is to investigate the performance of fine-tuning larger CNN models. In addition, finding the optimal fine-tuning depth, i.e. instead of fine-tuning all layers, some shallowers layers can be fixed while deeper layers can be fine-tuned, is an interesting area to explore.


\addtolength{\textheight}{-12cm}   








\bibliographystyle{plain}
\bibliography{bib}

\end{document}